\documentclass[twoside,]{article}
\usepackage{amsmath,amsfonts,bm,amssymb,xspace,bbm,mathtools}

\def\eqref#1{equation~\ref{#1}}
\def\1{\mathbbm{1}}

\def\vzero{{\bm{0}}}
\def\vone{{\bm{1}}}

\def\vtheta{{\bm{\theta}}}

\def\vb{{\bm{b}}}

\def\vg{{\bm{g}}}
\def\vh{{\bm{h}}}

\def\vl{{\bm{l}}}

\def\vw{{\bm{w}}}
\def\vx{{\bm{x}}}
\def\vy{{\bm{y}}}

\def\vepsilon{{\bm \epsilon}}

\def\vtheta{{\bm \theta}}

\def\vphi{{\bm \phi}}

\def\mA{{\bm{A}}}
\def\mB{{\bm{B}}}
\def\mC{{\bm{C}}}

\def\mE{{\bm{E}}}
\def\mF{{\bm{F}}}

\def\mH{{\bm{H}}}
\def\mI{{\bm{I}}}
\def\mJ{{\bm{J}}}

\def\mL{{\bm{L}}}

\def\mU{{\bm{U}}}

\def\mLambda{{\bm{\Lambda}}}

\DeclareMathAlphabet{\mathsfit}{\encodingdefault}{\sfdefault}{m}{sl}
\SetMathAlphabet{\mathsfit}{bold}{\encodingdefault}{\sfdefault}{bx}{n}

\def\emH{{H}}

\def\gK{\mathcal{K}}

\def\gT{\mathcal{T}}

\def\gV{\mathcal{V}}

\newcommand{\E}{\mathbb{E}}
\newcommand{\R}{\mathbb{R}}

\DeclareMathOperator*{\argmin}{argmin}

\makeatletter
\DeclareRobustCommand\onedot{\futurelet\@let@token\@onedot}
\def\@onedot{\ifx\@let@token.\else.\null\fi\xspace}

\def\eg{\emph{e.g}\onedot} 
\def\ie{\emph{i.e}\onedot}

\def\wrt{w.r.t\onedot} 
\def\rhs{r.h.s\onedot}
\def\lhs{l.h.s\onedot}

\makeatother

\providecommand{\norm}[1]{\lVert#1\rVert}
\DeclareMathOperator{\diag}{diag}

\newcommand{\stand}{\mathrm{d}}

\usepackage[accepted]{aistats2023}

\usepackage{hyperref}
\usepackage{url}
\usepackage{booktabs}
\usepackage{enumitem}
\usepackage[capitalize,noabbrev]{cleveref}
\usepackage{multirow}
\usepackage{xspace}

\usepackage{algorithm}
\usepackage{algpseudocode}

\newtheorem{theorem}{Theorem}
\newtheorem{remark}{Remark}

\usepackage{xcolor}
\definecolor{Blue}{RGB}{0,90,255}
\definecolor{Red}{RGB}{255,75,0}
\definecolor{Green}{RGB}{3,175,122}
\definecolor{WBlue}{RGB}{191,228,255}
\definecolor{WRed}{RGB}{255,202,191}
\definecolor{gray}{RGB}{127,127,127}

\newcommand{\nystrom}{Nystr\"om\xspace}

\usepackage[sortcites,
backend=biber,
hyperref=true,
style=authoryear,
language=auto,
maxbibnames=50,
maxcitenames=2,
natbib=true,
eprint=false]{biblatex}
\begin{document}

\runningauthor{Ryuichiro Hataya \& Makoto Yamada}
\twocolumn[

\aistatstitle{Nystr\"om Method for Accurate and Scalable Implicit Differentiation}

\aistatsauthor{Ryuichiro Hataya${}^{1,2}$ \And Makoto Yamada${}^{2,3,4}$}

\aistatsaddress{${}^{1}$RIKEN ADSP \And ${}^{2}$RIKEN AIP \And ${}^{3}$OIST \And ${}^{4}$Kyoto University} ]

\begin{abstract}
The essential difficulty of gradient-based bilevel optimization using implicit differentiation is to estimate the inverse Hessian vector product with respect to neural network parameters.
This paper proposes to tackle this problem by the \nystrom method and the Woodbury matrix identity, exploiting the low-rankness of the Hessian.
Compared to existing methods using iterative approximation, such as conjugate gradient and the Neumann series approximation, the proposed method avoids numerical instability and can be efficiently computed in matrix operations without iterations.
As a result, the proposed method works stably in various tasks and is faster than iterative approximations.
Throughout experiments including large-scale hyperparameter optimization and meta learning, we demonstrate that the \nystrom method consistently achieves comparable or even superior performance to other approaches.
The source code is available from \url{https://github.com/moskomule/hypergrad}.

\end{abstract}

\section{Introduction}

Bilevel optimization is an essential problem in machine learning, which includes hyperparameter optimization (HPO) \citep{hutter19} and meta learning \citep{hospedales2021meta}. 
This problem consists of an inner problem to minimize an inner objective $f(\vtheta, \vphi, \gT)$ on data $\gT$ with respect to parameters $\vtheta\in\R^p$ and an outer problem to minimize an outer objective $g(\vtheta, \vphi, \gV)$ on data $\gV$ with respect to hyper or meta parameters $\vphi\in\R^h$.
In the case of HPO, $f$ and $g$ correspond to a training loss function and a validation criterion. In contrast, in the case of meta learning, $f$ and $g$ correspond to meta-training and meta-testing objectives.

Typically in the deep learning literature, the bilevel optimization problem can be formulated as
\begin{align}
	&\min_\vphi g(\vtheta_T(\vphi), \vphi, \gV) \label{eq:bilevel_outer} \\
	\text{s.t.}&\quad\vtheta_t(\vphi)=\Theta(\vtheta_{t-1}(\vphi), \nabla_\vtheta f(\vtheta_{t-1}(\vphi), \vphi, \gT), \vphi), \label{eq:bilevel_inner}
\end{align}
\noindent where $\Theta$ is a gradient-based optimizer, such as SGD and Adam \citep{Kingma2015}, and $t=1,2,\dots,T$.
In some cases, the outer problem (\ref{eq:bilevel_outer}) can also be optimized by gradient-based optimization methods by using \emph{hypergradient} $\nabla_\vphi g$, in a similar way to the inner problem, which is expected to be more efficient and scalable than black-box counterparts.
Especially when combined with warm-start bilevel optimization that alternately updates outer parameters as \cref{eq:bilevel_outer} and inner parameters as \cref{eq:bilevel_inner} during training \citep{jaderberg2017,vicol22a}, the gradient-based approaches enjoy higher efficiency \citep{luketina16,lorraine20a}.

A straightforward approach to achieve this goal is to unroll the inner problem to back-propagate through \cref{eq:bilevel_inner} for hypergradient \citep{Finn2017b,grefenstette2019,Domke2012}. 
However, unrolling increases the memory cost as the number of inner optimization $T$ increases, which may also cause gradient vanishing/explosion \citep{antoniou2018}. 
Truncating the backward steps \citep{Shaban2019} may unburden these issues while sacrificing the quality of hypergradients.

Alternatively, approximating hypergradient using implicit differentiation is promising because it requires much less space complexity than the unrolling approach.
Exact implicit differentiation needs computationally demanding inverse Hessian vector product, which has been approximated by iterative methods such as conjugate gradient \citep{Pedregosa2016a,Rajeswaran2019} and the Neumann series approximation \citep{lorraine20a}. 
Thanks to their space efficiency, these methods can scale to large-scale problems \citep{lorraine20a,Hataya2022,li2021autobalance,zhang21s}, but such iterative approximations cost time complexities.
Furthermore, these methods need careful configuration tuning to avoid numerical instability caused by ill-conditioned Hessian or the norm of Hessian.

In this paper, we propose to use the \nystrom method to leverage the low-rank nature of Hessian matrices of neural networks and compute its inverse by the Woodbury matrix identity, inspired by recent works in quasi second-order optimization literature \citep{singh2020,singh2021nysnewton}.
Unlike the iterative approaches mentioned above, this approximation excludes iterative evaluations and can be computed instantly in matrix operations.
Additionally, the proposed method avoids numerical instability.
As a result, the \nystrom method is robust to configurations, and empirically compares favorably with existing approximation methods consistently on various tasks, from HPO to meta learning.
In addition, by using the recurrence of the Woodbury matrix identity, this approach can control the tradeoff between time and space complexities without losing accuracy according to one's computational resource.

In the remaining text, we introduce the proposed method in \cref{sec:method}. 
After reviewing related work in \cref{sec:related_work}, we analyze the approximation quality of the proposed method when Hessian is low-rank in \cref{sec:theoretical_analysis}. 
Then, \cref{sec:experiments} empirically demonstrates the effectiveness of the method from a synthetic logistic regression problem to a large-scale real-world data reweighting problem, and finally \cref{sec:conclusion} concludes this work.

\section{Method}\label{sec:method}

\subsection{Approximating Hypergradient by Implicit Differentiation}

In this paper, we focus on the methods to approximate hypergradients $\nabla_\vphi g$ by implicit differentiation so that the outer problem can also be efficiently optimized by gradient descent.
Specifically, if $\nabla_\vtheta f(\vtheta_T, \vphi)\approx \vzero$, then according to the implicit function theorem, we obtain
\begin{equation}\label{eq:implicit_differentiation}
	\frac{\stand g(\vtheta_T, \vphi)}{\stand \vphi}
	=-\frac{\partial g}{\partial \vtheta}
	\left(\frac{\partial^2 f}{\partial \vtheta^2}\right)^{-1}
	\frac{\partial^2 f}{\partial \vphi\partial\vtheta}
	+\frac{\partial g}{\partial \vphi},
\end{equation}
\noindent where, in the \rhs, $f=f(\vtheta_T, \vphi)$ and $g=g(\vtheta_T, \vphi)$.
Following the prior works \citep{Pedregosa2016a,Rajeswaran2019,lorraine20a}, we assume that this approximation holds after $T$ iterations of the inner optimization.
We also assume that factors in the \rhs of \cref{eq:implicit_differentiation} are available, \eg, $g$ is differentiable \wrt $\vtheta$. In some cases, such as  optimization of hyperparameters for regularization, $\nabla_\vphi g(\vtheta_T, \vphi)$ is always zero.

Still solving \cref{eq:implicit_differentiation} seems computationally intractable, as computing inverse Hessian $(\nabla_\vtheta^2f)^{-1}$ is computationally expensive, when the number of model parameters $p=\dim\vtheta$ is large.
Though early works compute inverse Hessian directly \citep{Larsen1996,Bengio2000}, especially for modern neural networks, just storing Hessian $\nabla_\vtheta^2f$ is already infeasible in practice.

To mitigate this issue, some approximations have been proposed.
\cite{Pedregosa2016a,Rajeswaran2019} used the conjugate gradient method \citep{Hestenes1952}, which iteratively solves a linear equation $\mA\vx=\vb$ to obtain $\vx=\mA^{-1}\vb$, where, in this case, $\mA=\nabla_\vtheta^2 f$ and $\vb=\nabla_\vtheta g$.
\cite{lorraine20a} adopted the Neumann series approximation, $\mA^{-1}=\alpha\sum_{i=1}^\infty (\mI-\alpha\mA)^i$, where $\mA$ is an invertible matrix that satisfies $\norm{\alpha A}\leq 1$ and $\alpha>0$ is a constant.
As these algorithms may take arbitrarily long iterations for convergence, their truncated versions are preferred in practice, which cut off the iterations at a predefined number of steps $l$.

Importantly, these methods do not require keeping actual Hessian but accessing it as Hessian vector product (HVP), which modern automatic differentiation tools \citep{pytorch,jax} can efficiently compute in $O(p)$ \citep{baydin2018}.
Because they consist of HVP and vector arithmetics, their time and space complexities are $O(lp+h)$ and $O(p+h)$ for the number of iterations $l$, where $p=\dim\vtheta$ and $h=\dim\vphi$. In the following discussion, we omit the complexity regarding $h$ for simplicity.

The downside of conjugate gradient and the Neumann series approximation may be their numerical instability.
Conjugate gradient needs a well-conditioned matrix for fast convergence \citep{saad2003iterative,golub2013matrix}, \ie, it works sub-optimally with ill-conditioned Hessian.
Its longer iterations accumulate numerical errors, and these errors typically need to be alleviated by pre-conditioning or reorthogonalization, which requires extra time and space complexities.
The Neumann series needs the matrix norm to be less than 1, and thus $\alpha$ needs to be carefully configured.

\subsection{\nystrom Approximation}

Different from these previous methods using iterative computation, we instead propose to use a low-rank approximation for approximated inverse Hessian vector product (IHVP).
Specifically, we propose to use the \nystrom method to obtain IHVP by leveraging the low-rank nature of Hessian matrix \citep{lecun2012efficient,karakida2019,Ghorbani19}.

We use the following $k$-rank approximation to the original $p$ dimensional Hessian matrix, where we assume $k\ll p$:
\begin{equation}
	\mH_k=\mH_{[:, K]}\mH_{[K, K]}^\dagger\mH_{[:, K]}^\top,
\end{equation}
\noindent where $K$ is a randomly selected index set of size $k$, $\mH_{[:, K]}\in\R^{p\times k}$ is a matrix extracted columns of $\mH$ corresponding to indices in $K$, and $\mH_{[K, K]}\in\R^{k\times k}$ is a matrix extracted rows of $\mH_{[:, K]}$ corresponding to indices in $K$.
$\mH_{[K, K]}^\dagger=\mU\mLambda^{-1}\mU^\top$ denotes the pseudo inverse, where $\mU$ and $\mLambda$ are eigenvectors and engenvalues of $\mH_{[K, K]}$.

Then, we use the Woodbury matrix identity for matrices $\mA, \mB, \mC$ such that
\begin{align}\label{eq:woodbury}
	&(\mA+\mC\mB\mC^\top)^{-1} \\
   =&\mA^{-1}-\mA^{-1}\mC(\mB^{-1}+\mC^\top\mA^{-1}\mC)^{-1}\mC^\top\mA^{-1} \notag
\end{align}
\noindent to obtain the inverse Hessian. Namely, we compute $(\mH_k+\rho\mI_p)^{-1}$, where $\rho>0$ is a small constant to improve numerical stability and $\mI_p$ is the $p$-dimensional identity matrix, to approximate $\mH^{-1}$ as follows:
\begin{align}\label{eq:full_inversion}
  & (\rho\mI_p+\mH_{[:, K]}\mH_{[K, K]}^\dagger\mH_{[:, K]}^\top)^{-1} \\
= & \frac{1}{\rho}\mI_p-\frac{1}{\rho^2}\mH_{[:, K]}\left(\mH_{[K,K]}+\frac{1}{\rho}\mH_{[:, K]}^\top\mH_{[:, K]}\right)^{-1}\mH_{[:, K]}^\top. \notag
\end{align}
Although this left-hand side requires the inversion of a $p\times p$ matrix, the right-hand side only needs the inversion of a $k\times k$ matrix. 
Because $k\ll p$, the computational burden of the \lhs is drastically reduced in the \rhs.
The use of Woodbury matrix identity as \cref{eq:full_inversion} is similar to the idea of \cite{singh2021nysnewton}, but our formulation is slightly efficient as it avoids unnecessary eigen decomposition.

The small constant $\rho$ in \cref{eq:full_inversion} makes a low-rank matrix $\mH_k$ invertible.
Additionally, it can also be regarded as being stemmed from a proximally regularized inner objective $\displaystyle f(\vtheta, \vphi)+\frac{\rho}{2}\norm{\vtheta-\vtheta'}$, where $\displaystyle\vtheta'\in\argmin_\vtheta f(\vtheta, \vphi)$ \citep{vicol22a}.
\begin{figure*}[t!]
	\centering
	\includegraphics[width=\linewidth]{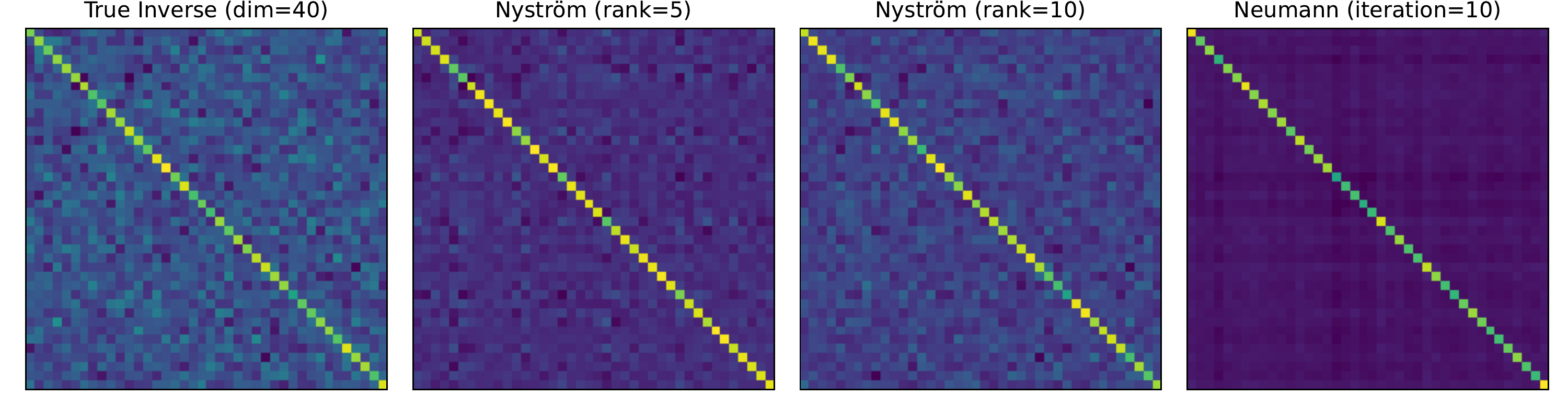}
    \vspace{-1.5\baselineskip}
	\caption{Comparison of inverse of a 40 dimensional matrix $\mA+\rho\mI$. $\mA$ is a rank 20 symmetric matrix, and $\rho$ is set to $0.1$. \nystrom method can approximate the true inverse accurately even in the rank 5 setting.}
	\label{fig:inverse_comparison}
\end{figure*}

We visualize the inverse of a low-rank matrix and its approximations in \cref{fig:inverse_comparison}. \nystrom method can approximate the true inverse efficiently and accurately. 
Because conjugate gradient cannot explicitly output inverse Hessian, we do not display its result here.

To sum up, the proposed method approximates the hypergradient by using a low-rank Hessian $\mH_k$ as
\begin{equation}
	\frac{\stand g(\vtheta_T, \vphi)}{\stand \vphi}
	\!\approx\!-\frac{\partial g}{\partial \vtheta}
	(\mH_k+\rho\mI_p)^{-1}
	\frac{\partial^2 f}{\partial \vphi\partial\vtheta}
	+\frac{\partial g}{\partial \vphi}.
\end{equation}

\subsection{Space-efficient Variant}

The \nystrom approximation is free from iterative algorithms, but it needs to store $k$ columns of the original Hessian matrix $\mH_{[:, K]}$ and compute the inverse of a $k\times k$ matrix.
As a result, its time and space complexities are $O(p+k^3)$ and $O(kp+k^2)$, but $k^3$ and $k^2$ are ignorable because usually $k\ll p$.

Some readers may worry about memory explosion when $k$ is relatively large.
Actually, this \nystrom approximation can be turned into an iterative algorithm that saves memory. 
Recall that the low-rank matrix can be decomposed as follows
\begin{equation}
	\mH_k=\mH_{[:, K]}\mH_{[K, K]}^\dagger\mH_{[:, K]}^\top=\sum_{i\in K}\frac{1}{\lambda_i}\vl_i\vl_i^\top,
\end{equation}
\noindent where $\lambda_i\in\R$ is the $i$th value of $\mLambda$ and $\vl_i=(\mH_{[:,K]}\mU)_{[:,i]}\in\R^p$.
Then, we can iteratively compute the inverse of $\mH_k+\rho\mI_p$ by the Woodbury matrix identity (\cref{eq:woodbury}) as
\begin{align}\label{eq:rank1_inversion}
	\hat{\mH}_{i+1}
  &=\hat{\mH}_{i}-\frac{\hat{\mH}_{i}\vl_i\vl_i^\top\hat{\mH}_{i}}{\lambda_i+\vl_i^\top\hat{\mH}_{i}\vl_i}, \\
  \text{where} & \quad \hat{\mH}_{0}
  =\frac{1}{\rho}\mI_p,\notag \\
  & \quad \hat{\mH}_{k}=(\mH_k+\rho\mI_p)^{-1}, \notag
\end{align}
\noindent for $i=0,1,\dots,k-1$. This variant needs $O(k^2p)$ time complexity and $O(p)$ space complexity like iterative algorithms. \cite{singh2020} proposed a dynamical algorithm for a similar problem to compute the inverse of the Fisher information matrix (FIM).

\subsection{Controlling the Cost Tradeoff}

Furthermore, by chunking $\mH_{[:,K]}$ into thinner matrices of width $\kappa\in(1, k)$ and applying the Woodbury matrix identity iteratively as \cref{alg}, $(\mH_k+\rho\mI_p)^{-1}$ can be obtained in a hybrid manner with less memory footprint, \ie, $O(\kappa p)$, than \cref{eq:full_inversion} and faster, \ie, $O(\{k/\kappa\}^2 p)$, than \cref{eq:rank1_inversion}.
In other words, our method allows users to dynamically control the necessary tradeoff between time and space complexities for given accuracy, which is a unique characteristic of our proposed method.
See \cref{tab:complexity} for comparison with other methods.

Notice that for any $\kappa$, the computational result is equivalent to each other up to machine precision.
Thus, in the remaining paper, we use \cref{eq:full_inversion}, where $\kappa=k$, without otherwise specified.

\begin{table*}[t]
	\centering
	\caption{Comparison of time and space complexity. $p$ denotes the number of model parameters. $l$ is the number of iterations of the algorithms, $k$ is the rank of low-rank Hessian. The \nystrom method allows users to control the complexities by choosing $\kappa\in\{1, 2, \dots, k\}$}.
	
	\begin{tabular}{lcc}
		\toprule
		Approximation      &  Time Complexity  &  Space Complexity \\
		\midrule
		Conjugate gradient \citep{Rajeswaran2019} &  $O(lp)$          &  $O(p)$           \\
		Neumann series     &  $O(lp)$          &  $O(p)$           \\
		\nystrom method (ours)    &  $O((k/\kappa)^2 p)$&  $O(\kappa p)$   \\
		\bottomrule
	\end{tabular}
	
	\label{tab:complexity}
\end{table*}
\begin{algorithm}[t]
	\caption{Algorithm of the proposed method}\label{alg}
	\begin{algorithmic}
	\Require \\
	\begin{description}
		\item[$\kappa$:] control parameter of computational cost
		\item[$K$:] randomly selected index set, where $\#K=k$
		\item[$\mH_{[:, K]}$:] a column matrix of $\mH$ corresponding to $K$
		\item[$\mU, \mLambda$:] eigen decomposition of $\mH_{[K, K]}=\mU\mLambda\mU^\top$
	\end{description}
	\\
	\State Partition $K$ in to size $\kappa$ subsets $\gK$
	\State $\hat{\mH}=\frac{1}{\rho}\mI_p$
	\For{$K'$ in $\gK$}
		\State $\mL\gets(\mH_{[:, K]}\mU)_{[:, K']}$
		\State $\mJ\gets \mLambda_{[K', K']}$
		\State $\hat{\mH}\gets \hat{\mH}-\hat{\mH}\mL(\mJ+\mL^\top\hat{\mH}\mL)^{-1}\mL^\top\hat{\mH}$
	\EndFor
	\State Return $\hat{\mH}$, equivalent to $(\mH_k+\rho\mI_p)^{-1}$
	\end{algorithmic}
\end{algorithm}
\subsection{Limitations}

The proposed method cannot straightforwardly optimize outer parameters $\vphi$ that do not directly affect the training loss, inheriting the limitation of the methods to approximate hypergradient by implicit differentiation \citep{lorraine20a}. 
Such parameters include a learning rate of an optimizer of the inner problem, which needs to be carefully tuned in deep learning research \citep{schmidt21a}.
We may need to rely on unrolling approaches for this problem \citep{li2017learning,Andrychowicz2016,grefenstette2019}.
Additionally, the method does not directly applicable to non-smooth problems as other gradient-based methods, part of which could be alleviated by smoothing the problem or using sub-gradients and sub-Hessians.

\section{Related Work}\label{sec:related_work}

\subsection{Gradient-based Hyperparameter Optimization and Meta Learning}

The development of automatic differentiation \citep{baydin2018,jax,pytorch} has encouraged the active research of gradient-based HPO and meta learning, where the outer problem of a bilevel problem (\cref{eq:bilevel_outer}) is also optimized by gradient descent using hypergradient \citep{Franceschi2018,franceschi2021}.

One way to compute hypergradients is to backpropagate through the unrolled inner optimization (\cref{eq:bilevel_inner}) \citep{Finn2017b,grefenstette2019,Domke2012}. 
Except for the special cases, where a specific inner optimization algorithm \citep{Maclaurin2015} or forward-mode automatic differentiation \citep{franceschi17a} can be used, this approach suffers from space complexity as the inner optimization step $T$ increases.

Another approach is to approximate hypergradient using the implicit differentiation as \cref{eq:implicit_differentiation} with less space complexity \citep{Bengio2000,Pedregosa2016a,Rajeswaran2019,lorraine20a}.
Although \cref{eq:implicit_differentiation} includes inverse Hessian, which is infeasible to compute for modern neural networks, truncated solutions of conjugate gradient \citep{Pedregosa2016a,Rajeswaran2019} and the Neumann series approximation \citep{lorraine20a} have been adopted to approximate this term efficiently.
Other solvers for linear systems, such as GMRES \citep{gmres}, can also be used \citep{blondel2021}.
Our work is in line with these works, but compared to these methods using generic techniques for matrix inverse, the proposed method exploits the low-rankness of Hessian of neural networks.

\subsection{Inverse Hessian Approximation}

The application of inverse Hessian approximation is not limited to the computation of hypergradient.
It has been a key element in estimation of influence function \citep{Koh2017}, backpropagation through long recurrence \citep{Liao2018}, and network pruning \citep{hassibi1992second,singh2020}.

The inverse of Hessian or an FIM is also indispensable in (quasi) second-order optimization \citep{martens2010deep} and natural gradient descent \citep{amari98}. Thus, its estimation has been studied for a long time. 
For example, LBFGS employs past gradients and updates \citep{liu1989limited}, and KFAC adopts block-diagonal approximation of FIM \citep{martens15} for efficient approximation of large matrix inversion.

The Hessians and FIMs of neural networks have low-rank structures, as most of their eigenvalues are nearly zero \citep{lecun2012efficient,karakida2019,Ghorbani19}. 
Exploiting this nature, inverse FIM \citep{frantar2021} or inverse Hessian \citep{singh2021nysnewton} are computed using the Woodbury identity in the literature of quasi second-order optimization. Especially, the latter used the \nystrom method.
Although these approaches are technically similar to ours, they are in a different context.

\section{Theoretical Analysis}\label{sec:theoretical_analysis}

This section theoretically shows that the proposed approach can efficiently approximate the true hypergradient.
\begin{theorem}\label{thm:hypergrad}
	Suppose $\mH$ is a positive semidefinite. Let $\vh^\star$ and $\vh$ be hypergradients using the true inverse Hessian $(\mH+\rho\mI_p)^{-1}$ (\rhs of \cref{eq:implicit_differentiation}) and the \nystrom method $(\mH_k+\rho\mI_p)^{-1}$ (\rhs of \cref{eq:full_inversion}), and $\vg=\nabla_\vtheta g(\vtheta_T, \vphi)$, $\mF=\nabla_\vphi\nabla_\vtheta f(\vtheta_T, \vphi)$. Then, the accuracy of approximated hypergradient is bounded
	\begin{equation}\label{eq:hypergradient_error}
		\norm{\vh^\star-\vh}_2 \leq \norm{\vg}_2\norm{\mF}_\mathrm{op}\left(\frac{1}{\rho}\frac{\norm{\mH-\mH_k}_\mathrm{op}}{\rho+\norm{\mH-\mH_k}_\mathrm{op}}\right).
	\end{equation}
\end{theorem}

$\norm{\cdot}_2$ and $\norm{\cdot}_\mathrm{op}$ denote L2 norm and operator norm, respectively.

This theorem is based on \citep{frangella2021}. See the supplemental material for the derivation.

When considering neural networks, because the training objective $f$ is not convex \wrt $\vtheta$, its Hessian $\mH$ is not always positive semi-definite.
However, \cite{Ghorbani19} empirically demonstrated that most negative eigenvalues disappeared even after a few iterations of training, indicating that we may assume that $\mH+\rho \mI$ for $\rho>0$ is positive semi-definite in practice. Also importantly, $\norm{\mH-\mH_k}_\mathrm{op}$ is bounded.

\begin{remark}[Theorem 3 in \cite{drineas05a}]
	Let $\bar{\mH}_k$ be the best $k$-rank approximation of $\mH$. If $O(k/\epsilon^4)$ columns are selected for $\epsilon>0$ so that the $i$th column is chosen proportional to $\emH_{i,i}^2$, then,
	\begin{equation}
		\E[\norm{\mH-\mH_k}_\mathrm{op}]\leq \norm{\mH-\bar{\mH}_k}_\mathrm{op} + \epsilon \sum_{i=1}^{p} \emH_{i,i}^2.
	\end{equation}	
	Especially if $\mH$ is  a rank $k$ matrix, then
	\begin{equation}\label{eq:hessian_error}
	    \E[\norm{\mH-\mH_k}_\mathrm{op}]\leq \epsilon \sum_{i=1}^{p} \emH_{i,i}^2
	\end{equation}
\end{remark}

Because Hessian of a trained neural network can be regarded as low rank \citep{lecun2012efficient,karakida2019,Ghorbani19}, we may expect that \cref{eq:hessian_error} holds.

\Cref{eq:hypergradient_error} indicates that an approximated hypergradient converges to the true hypergradient as $\mH_k$ approaches to $\mH$.
This differs from truncated iterative approximations, such as conjugate gradient, where their expected solutions never converge to the true one for a small number of iterations.

\section{Experiments}\label{sec:experiments}

In this section, we empirically demonstrate the effectiveness of the proposed method.

\subsection*{Experimental Setups}

We implemented models and algorithms using \texttt{PyTorch} v1.12 \citep{pytorch} and its accompanying \texttt{functorch} \citep{functorch2021}.
The reference code is available from \url{https://github.com/moskomule/hypergrad}. 
Experiments were conducted on a single NVIDIA A100 GPU with CUDA 11.3.
The implementations of conjugate gradient and the Neumann series approximation algorithms were adopted from \texttt{betty} v0.1.1 \citep{choe2022betty}.

In the following experiments, the \nystrom method was implemented according to \cref{eq:full_inversion}, that is, the time-efficient variant otherwise spcified.
Using ReLU as an activation function leads some columns of Hessian to zero vectors, and then the inversion in \cref{eq:full_inversion} fails. To circumvent this problem, we replaced ReLU with leaky ReLU:  $\text{LR}(x)=\max(0, x)+0.01\times\min(0, x)$.

\subsection{Optimizing Weight-decay of Linear Regression}

We first showcase the ability of the \nystrom approximation by optimizing weight-decay parameters for each parameter of a linear regression model using synthetic data. 
For $D$ dimensional data, inner parameters $\vtheta\in\R^D$ and outer parameters $\vphi\in\R^D$ are optimized.
The inner problem is $f(\vtheta,\vphi)=\ell(\vtheta^\top\vx, y)+\vtheta^\top \diag(\vphi)\vtheta$, for an input $\vx$ and its label $y$, where $\ell$ is binary cross entropy loss. Each input $\vx$ is sampled from a standard normal distribution, and its label is defined as $y=\vw^{\ast\top}\vx+\vepsilon>0$, where $\vw^\ast\in\R^D$ is a constant vector and $\vepsilon\in\R^D$ is a noise vector. 
This inner problem is optimized by SGD with a learning rate of 0.1, and the inner parameters are reset every 100 iteration.
The outer problem is to minimize validation loss by SGD with a learning rate of 1.0 and a momentum of 0.9.
The outer parameters $\vphi$, initialized to $\vone$, are updated after every 100 inner parameter update.
We set $D=100$ and used 500 data points for both inner and outer optimization.

\begin{figure}[t]
	\centering
	\includegraphics[width=\linewidth]{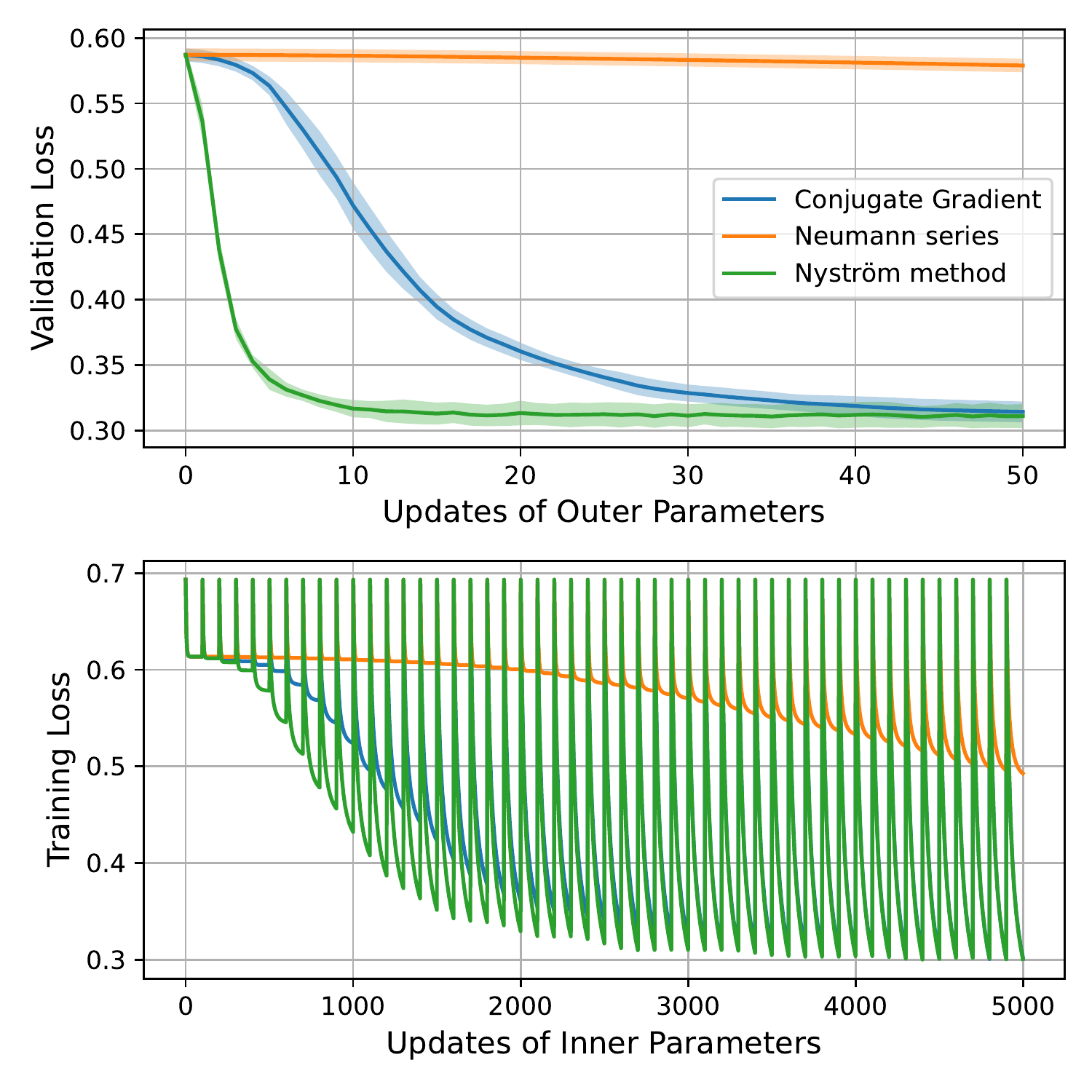}
    \vspace{-1.5\baselineskip}
	\caption{Optimization of weight decay parameters to each model parameter in logistic regression. The top figure shows the \emph{validation} loss curve of the outer problem, and the bottom figure shows the \emph{training} loss curves of the inner problem, optimized in 100 iterations.}
	\label{fig:logistic_regression}
\end{figure}

\Cref{fig:logistic_regression} (top) shows validation loss curves, comparing approximated implicit differentiation methods, conjugate gradient, and the Neumann series approximation, with our proposed method.
For the conjugate gradient method and the Neumann series approximation, we set the number of iterations $l$ to $5$, following \cite{Rajeswaran2019}. 
Accordingly, we set the rank of the \nystrom method to $5$.
As can be seen, the \nystrom method can optimize the weight-decay parameters faster than other methods.
\Cref{fig:logistic_regression} (bottom) displays training loss curves. 
Because the inner parameters are reset when the outer parameters are updated, training loss values at inner-parameter reset moments are high (around $0.7$).
As the outer optimization proceeds, the inner parameters, particularly those of the \nystrom method, quickly decreases the training loss during each inner optimization period.

\begin{figure}[t]
	\centering
	\includegraphics[width=\linewidth]{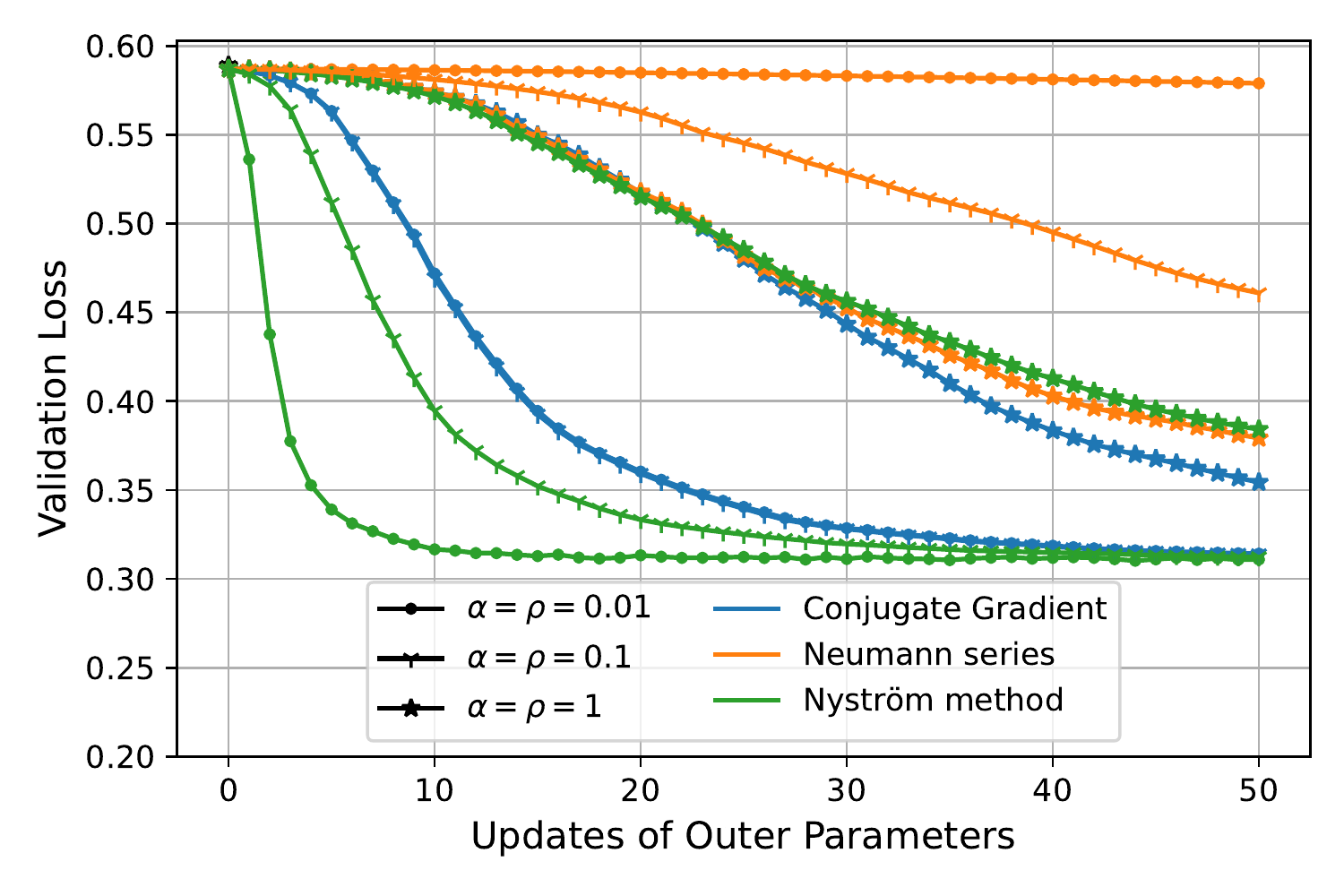}
    \vspace{-1.5\baselineskip}
	\caption{Validation loss curves of implicit differentiation methods with different configurations.}
	\label{fig:logistic_regression_params}
\end{figure}

For the experiments in \cref{fig:logistic_regression}, the ``learning rate'' parameter $\alpha$ of conjugate gradient and the Neumann series approximation was set to $0.01$, and $\rho$ of the \nystrom method was set to $0.01$.
We compare other choices of $\alpha$ in $\{0.01, 0.1, 1.0\}$ in \cref{fig:logistic_regression_params}.
Accordingly, we try other values of $\rho$ in $\{0.01, 0.1, 1.0\}$.
The results indicate that the \nystrom method surpasses others in most cases and show robustness to the choice of $\rho$.
We will revisit the robustness of the \nystrom method later in \cref{sub:robustness}.
These experimental results were averaged over five runs of different random seeds.

\subsection{Dataset Distillation}

Dataset distillation is a task to optimize a small synthesized training dataset $\gT_\vphi=\{\vphi_1, \vphi_2, \dots, \vphi_C\}$ parametrerized by $\vphi$ so that validation loss on real data is minimized \citep{wang2018}.
We used MNIST dataset \citep{lecun2010mnist} and a LeNet-like CNN, and followed the fixed-known initialization setting that CNN weights, \ie, the inner parameters, are reset every 100 model parameter update. 
As MNIST is a 10-class dataset, we set $C=50$, so each class has 5 distilled images. 
Each $\vphi_i\in\gT_\vphi$ has an equal size to an MNIST image.
We used fixed learning rates for inner and outer optimization to simplify the problem.
Namely, the inner problem is optimized by SGD with a learning rate of $0.01$, while the outer problem is optimized with an Adam optimizer with a learning rate of $1.0\times10^{-3}$.

The test accuracy after 5,000 outer parameter updates is reported in \cref{tab:dataset_disstillation_test}. 
These results were averaged over five runs.
We set $\alpha=\rho=0.01$ and $l=k=10$.

The \nystrom method yields comparable performance to the Neumann series approximation.
However, despite our best efforts to select appropriate values of $\alpha\in\{0.01, 0.1, 1.0\}$ and $l\in\{5, 10, 20\}$ based on the validation performance on a 10\% split of training data, the conjugate gradient method failed to learn this task.
This failure may be attributed to ill-conditioned Hessian.

\begin{table}[t]
    \centering
    \caption{Test accuracy of the dataset-distillation task on the MNIST dataset. The \nystrom method shows better performance than others.}
    
    \begin{tabular}{ccc}
         \toprule
         Conjugate gradient & Neumann series & \nystrom method \\ \midrule
         $0.17\pm0.04$      & $0.47\pm0.03$  & $0.49\pm0.04$ \\
         \bottomrule
    \end{tabular}
    
    \label{tab:dataset_disstillation_test}
\end{table}

\subsection{Gradient-based Meta Learning}

MAML is a typical method of gradient-based meta learning, where the inner problem learns to adapt to a given problem while the outer problem aims to find good parameters that adapts quickly to new tasks \citep{Finn2017b}. 
Among its variants, iMAML uses implicit differentiation to compute hypergradient, achieving better memory efficiency \citep{Rajeswaran2019}.
Although the original iMAML adopts conjugate gradient to obtain IHVP, this choice can be replaced with the Neumann series approximation and the \nystrom method.

We compared such backends using few-shot image classification, where models are learned to classify images only from few examples \citep{fei2006one,lake2011one,ravi2017optimization}, on the Omniglot dataset \citep{lake2015human} with a VGG-like CNN, following \citep{Rajeswaran2019,antoniou2018}.
We set $k=l=10$, $\alpha=\rho=0.01$. 
The inner problem is to optimize model parameters by SGD with a learning rate of 0.1 in 10 steps, and the outer problem is to update the initial model parameters by Adam with a learning rate of $1.0\times 10^{-3}$.

\Cref{tab:meta_learning} shows the averaged accuracy on the test tasks over three runs after training on $1.6\times 10^6$ tasks.
As can be seen, the \nystrom method achieved comparable results with iMAML using conjugate gradient both in the 1-shot and 5-shot settings.

\begin{table}[t]
    \centering
    \caption{Test accuracy of the meta learning task on the Omniglot dataset. The \nystrom method shows comparable performance to conjugate gradient.}
    \begin{tabular}{lcc}
        \toprule
        Task               &  1-shot          &  5-shot       \\
        \midrule
        Conjugate gradient &  $0.96\pm0.00$   &  $0.98\pm0.00$      \\
        Neumann series     &  $0.91\pm0.00$            &  $0.97\pm0.00$       \\
        \nystrom method    &  $0.95\pm0.00$            &  $0.98\pm0.00$        \\
        \bottomrule
    \end{tabular}
    \label{tab:meta_learning}
\end{table}

\subsection{Data Reweighting}

Data reweighting is a task to learn to weight a loss value to each example, which aims to alleviate the effect of class imbalance and label noise \citep{shu2019,li2021autobalance}. 
Its inner problem can be formulated as $f(\vtheta, \vphi)=\ell(\nu_\vtheta(\vx), \vy) \cdot \mu_\vphi(\ell(\nu_\vtheta(\vx), \vy))$, where $\ell$ is cross-entropy loss, $\nu_\vtheta$ is a model, and $\mu_\vphi$ is a neural network to weight samples.
The outer problem is to update $\vphi$ to minimize validation loss on balanced validation data.
The inner parameters are not reset when the outer parameters are updated.

We adopted long-tailed CIFAR-10 datasets \citep{cui2019class}, which simulate class imbalance at several degrees, WideResNet 28-10 \citep{Zagoruyko2016} as $\nu_\vtheta$, which has approximately $3.6\times 10^7$ parameters, and a two-layer MLP with a hidden dimension of $100$ as $\mu_\vphi$. 
The inner problem is optimized by SGD with a learning date of 0.1, momentum of 0.9, and weight decay of $5.0\times10^{-4}$, and the outer problem is optimized with an Adam optimizer with a learning rate of  $1.0\times10^{-5}$ on 2\% split of training data, following \cite{shu2019}.
We set $\alpha=\rho=0.01$ and $l=k=10$.

\Cref{tab:data_reweighting} shows the averaged test accuracy over three runs after $1.5\times 10^4$ inner updates and $1.5\times 10^3$ outer updates.
Again, the \nystrom method consistently yielded matching or better performance to other methods and outperformed the baseline.

\begin{table*}[h!]
    \centering
    \caption{Test accuracy of the data reweighting task on the long-tailed CIFAR-10 datasets. Baseline indicates training without outer optimization. The \nystrom method achieves consistently favorable results.}
    \begin{tabular}{lccc}
        \toprule
        Imbalanced factor   &  200            &  100        &  50              \\
        \midrule
        Baseline           &   $0.62\pm0.06$  &  $0.67\pm0.13$& $0.74\pm0.08$             \\
        Conjugate gradient &   $0.63\pm0.06$  &  $0.70\pm0.05$& $0.78\pm0.02$    \\
        Neumann series     &   $0.60\pm0.09$  &  $0.73\pm0.01$& $0.79\pm0.01$    \\
        \nystrom method    &   $0.66\pm0.02$  &  $0.73\pm0.02$& $0.79\pm0.01$        \\
        \bottomrule
    \end{tabular}
    
    \label{tab:data_reweighting}
\end{table*}

\subsection*{Runtime Speed and Memory Consumption}

\Cref{tab:runtime_stats} compares speed and peak GPU memory consumption to compute hypergradients on the data reweighting task (averaged over 10 runs). 
Because WideResNet 28-10 caused out of memory when $k=20$ with the time-efficient \nystrom method, we instead used relatively smaller WideResNet 28-2, which has $1.5\times 10^6$ parameters. The reported values were measured after 10 iterations of warmup.

As shown in \cref{tab:complexity}, the time complexity of iterative algorithms, conjugate gradient and the Neumann series, depends on $l$, whereas that of the \nystrom method is independent of $k$. 
As a result, the runtime speed of iterative algorithms slowdowns as the approximation quality $l$ increases, while the deceleration of the time-efficient \nystrom method is marginal.
On the other hand, the space complexity of the iterative algorithms is constant of $l$, which is reflected in the results.  
In contrast, that of the time-efficient \nystrom method relies on $k$, which can also be observed from the linear growth of the actual memory consumption.

\cref{tab:runtime_stats} also presents the results of the space-efficient variant of the \nystrom method, where $\kappa=1$.
Its memory consumption is constant, while the speed is almost quadratic to $k$, demonstrating the controllability of the tradeoff between speed and memory consumption as expected in \cref{tab:complexity}.

\begin{table*}[t]
    \centering
    \caption{Average runtime speed and peak memory consumption for hypergradient computation in data reweighting task over 10 runs.}
    
    \begin{tabular}{llcc}
    \toprule
                       &           & Speed (s)        & Peak GPU Memory Consumption (GB) \\
    \midrule
    Conjugate gradient \citep{Pedregosa2016a} & $l=5$     & 0.44             & 2.46                   \\
                       & $l=10$    & 0.83             & 2.46                   \\
                       & $l=20$    & 1.68             & 2.46                   \\
    \midrule
    Neumann series \citep{lorraine20a}    & $l=5$     & 0.40             & 2.39                       \\
                       & $l=10$    & 0.75             & 2.39                        \\
                       & $l=20$    & 1.48             & 2.39                    \\
    \midrule
    \nystrom method (ours)   & $k=5$     & 0.24             & 4.66                          \\
    (time efficient)   & $k=10$    & 0.33             & 8.15                          \\
                       & $k=20$    & 0.54             & 15.1                          \\
    \midrule
    \nystrom method (ours)   & $k=5$     & 3.11       & 1.94                       \\
    (space efficient)  & $k=10$    & 10.7             & 1.94                          \\
                       & $k=20$    & 41.0              & 1.94                          \\
    \bottomrule
    \end{tabular}
    
    \label{tab:runtime_stats}
\end{table*}

\subsection*{Robustness of the \nystrom method}\label{sub:robustness}

\begin{figure}[t]
    \centering
    \includegraphics[width=\linewidth]{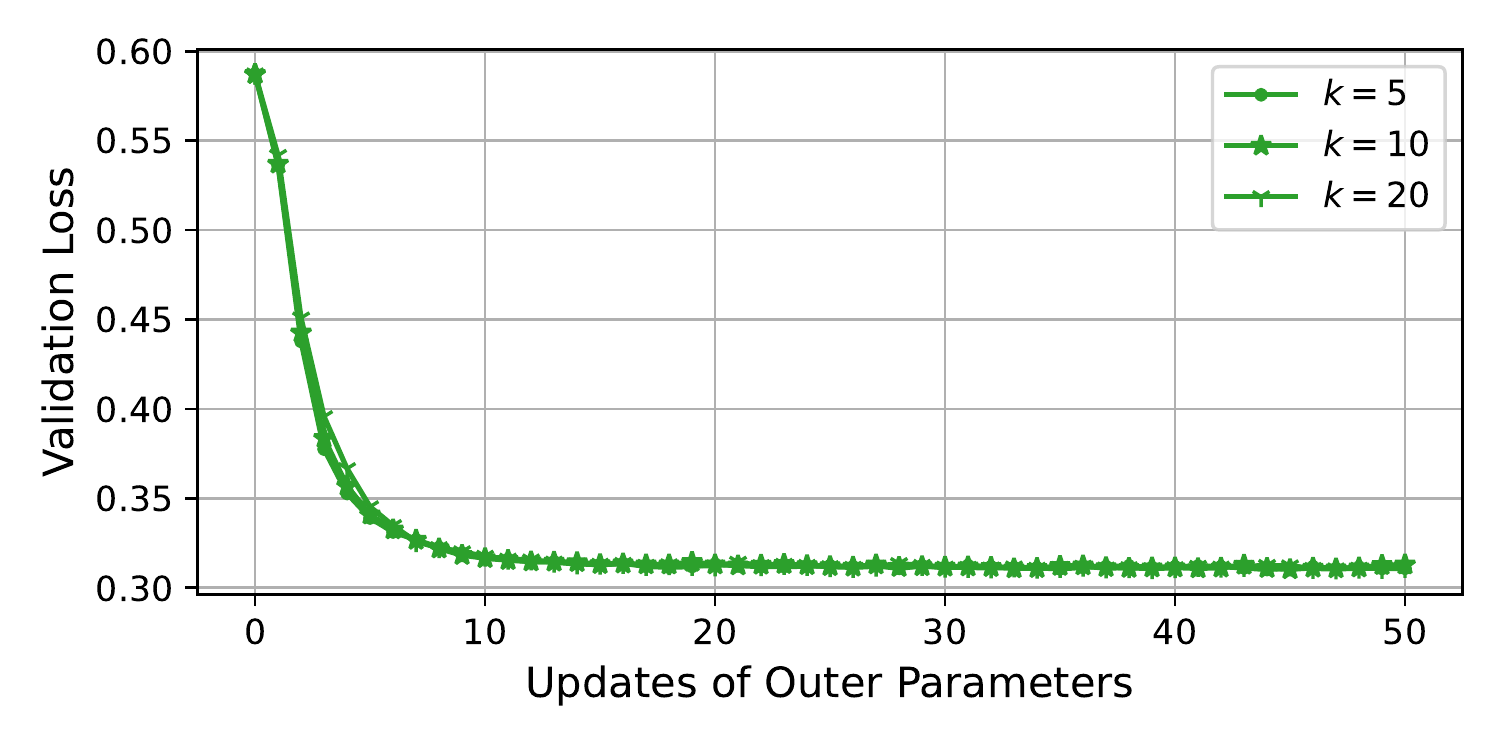}
    \vspace{-1.5\baselineskip}
    \caption{The effect of $k$ when $\rho=0.01$ in weight-decay optimization of logistic regression.}
    \label{fig:robustness_logistic}
\end{figure}

The \nystrom method has two parameters $\rho$ for numerical stability and $k$ for the matrix rank.
\Cref{tab:rho_k} shows the effect of these configurations on the data reweighting task using WideResNet 28-2 on the long-tailed CIFAR-10 of the imbalanced factor of 50 over $\rho\in\{0.01, 0.1, 1.0\}$ and $k\in\{5, 10, 20\}$.
The results differ only marginally, \ie, the proposed method is robust to the choice of configurations, which is a favorable property in practical applications.
\Cref{fig:robustness_logistic} compares the validation curves in weight-decay optimization of logistic regression using the \nystrom method with different $k$.
Again, the differences of curves among configurations are marginal, emphasizing the robustness of the \nystrom method.

These results suggest that $k=5$ may be sufficient for practically sized problems, which is faster than other methods, while consuming only twice memory (\cref{tab:runtime_stats}).
Also notice that, throughout various experiments including HPO and meta learning, the proposed method successfully and consistently works, different from other methods that failed at some tasks.
This indicates that the \nystrom method may also be robust to the types of problems.
These properties are appealing for practical use cases, that is, the \nystrom method may need minimum efforts for ``\emph{hyper}-hyperparameter optimization.''

\begin{table}[t]
    \centering
    \caption{The effect of $\rho$ and $k$ of the \nystrom method on the data reweighting task. Test accuracy is reported. The baseline without outer optimization yields test accuracy of $0.75\pm0.03$. These results indicate the robustness of the proposed method to configurations.}
    \begin{tabular}{ccccc}
    \toprule
         &   &      &  $\rho$    &      \\
         &   &  0.01&   0.1       &  1.0 \\
        \cmidrule{3-5}
         &    5 & $0.79\pm0.01$ & $0.78\pm0.01$ & $0.79\pm0.01$    \\
    $k$  &    10& $0.79\pm0.01$ & $0.78\pm0.01$ & $0.78\pm0.01$   \\
         &    20& $0.78\pm0.02$ & $0.78\pm0.01$ & $0.79\pm0.01$   \\
         \bottomrule
    \end{tabular}
    
    \label{tab:rho_k}
\end{table}

\section{Conclusion and Discussion}\label{sec:conclusion}

This paper introduced an approximated implicit differentiation method for gradient-based bilevel optimization using the \nystrom method. 
The key idea was to exploit the low-rank property of Hessian of neural networks by the \nystrom method and use the Woodbury matrix identity for fast and accurate computation of inverse Hessian vector product in hypergradient.
The proposed method scaled to large-scale problems and was applicable to hyperparameter optimization and meta learning.
Empirically, the approach was robust to configurations and about two times faster than iterative approximation methods.

Although hyperparameter optimization is crucial in machine learning, especially in deep learning, traditional hyperparameter optimization is costly and emits a substantial amount of CO2 \citep{strubell2020energy}.
Contrarily, gradient-based hyperparameter optimization is efficient and may help alleviate this issue.
Since the proposed method is fast, scalable, robust, and applicable to a wide range of tasks, it may provide a reliable way for researchers and practitioners to introduce efficient bilevel optimization.

\section*{Acknowledgments}

We thank anonymous reviewers for constructive comments to improve the manuscript.
R.H. was supported by JST, ACT-X Grant Number JPMJAX210H, and M.Y. was supported by MEXT KAKENHI Grant Number 20H04243.
We used computational resources of ``mdx: a platform for the data-driven future.''

% \newpage
\printbibliography

\newpage
\appendix
\onecolumn

\section*{Supplemental Material of ``Nystr\"om Method for Accurate and Scalable Implicit Differentiation''}

\section{Proof of \cref{thm:hypergrad}}

For positive semi-definite matrices $\mH$ and $\mH_k$, we have 

\begin{align*}
\|\vh^\ast - \vh\|_2 &\leq \left \|-\frac{\partial g(\vtheta_T, \vphi)}{\partial \vtheta}
	(\mH + \rho \mI_p)^{-1}
	\frac{\partial^2 f}{\partial \vphi\partial\vtheta}
	+\frac{\partial g(\vtheta_T, \vphi)}{\partial \vphi} - \left(-\frac{\partial g(\vtheta_T, \vphi)}{\partial \vtheta}
	(\mH_k+\rho\mI_p)^{-1}
	\frac{\partial^2 f}{\partial \vphi\partial\vtheta}
	+\frac{\partial g(\vtheta_T, \vphi)}{\partial \vphi}\right)\right\|_2\\
  &= \left\|-\frac{\partial g(\vtheta_T, \vphi)}{\partial \vtheta}
	\left\{(\mH + \rho \mI_p)^{-1} -(\mH_k+\rho\mI_p)^{-1}\right\}
	\frac{\partial^2 f}{\partial \vphi\partial\vtheta}\right \|_2\\
  &\leq \left\|\frac{\partial g(\vtheta_T, \vphi)}{\partial \vtheta}\right \|_2 \left \| \{(\mH + \rho \mI_p)^{-1} -(\mH_k+\rho\mI_p)^{-1}\}
	\frac{\partial^2 f}{\partial \vphi\partial\vtheta} \right\|_{\text{op}}\\
  &\leq \left\|\frac{\partial g(\vtheta_T, \vphi)}{\partial \vtheta}\right \|_2 \left \| (\mH + \rho \mI_p)^{-1} -(\mH_k+\rho\mI_p)^{-1} \right \|_{\text{op}}
	\left \|\frac{\partial^2 f}{\partial \vphi\partial\vtheta} \right\|_{\text{op}}\\
 & \leq \left\|\frac{\partial g(\vtheta_T, \vphi)}{\partial \vtheta}\right \|_2 
	\left \|\frac{\partial^2 f}{\partial \vphi\partial\vtheta} \right\|_{\text{op}} \left(\frac{1}{\rho} \frac{\|\mE\|_{\text{op}}}{\|\mE\|_{\text{op}} + \rho} \right),
 \end{align*}
 where $\mE = \mH - \mH_{k}$. In the final step, we used proposition 3.1 of \citet{frangella2021}.

\end{document}